\pdfoutput=1
\documentclass[11pt]{article}
\usepackage[utf8]{inputenc}
\usepackage[T1]{fontenc}
\usepackage{times}
\usepackage{latexsym}
\usepackage{inconsolata}

\usepackage{amsmath, amssymb, mathtools, amsthm}
\DeclareMathOperator*{\argmax}{argmax}

\usepackage{graphicx}
\usepackage{caption}
\usepackage{subcaption}
\usepackage{booktabs}
\usepackage{multirow}
\usepackage{array}
\usepackage{tabularx}
\usepackage{longtable}
\usepackage{colortbl}
\usepackage{arydshln} % 虚线表格线
\newcolumntype{R}[1]{>{\raggedleft\arraybackslash}p{#1}} % 定义右对齐列类型

\usepackage{microtype}
\usepackage{xspace}
\usepackage{xcolor}
\usepackage{wrapfig}
\usepackage{float}
\usepackage{adjustbox}
\usepackage{enumitem}
\usepackage{placeins}

\usepackage{hyperref}
\usepackage{url}
\usepackage[capitalize,noabbrev]{cleveref}

\usepackage{algorithm}
\usepackage{algorithmic} % 或 \usepackage{algorithmicx}, \usepackage{algpseudocode}

\usepackage{listings}

\usepackage{pifont}

\newcommand{\vpara}[1]{\vspace{0.05in}\noindent \textbf{#1 }}

\usepackage[final]{acl}

\title{From Objectives to Questions: A Planning-based Framework for Educational Mathematical Question Generation}

\author{Cheng Cheng$^{1,2}$, Zhenya Huang$^{1,2,3}$\thanks{Corresponding author},Guanhao Zhao$^{1,2}$,  Yuxiang Guo$^{1,2}$,  Xin Lin$^{1,2}$, \\ {\bf Jinze Wu$^{2,4}$, Xin Li$^{2,4}$,  Shijin Wang$^{2,4}$}\\
$^1$ University of Science and Technology of China \\ 
$^2$ State Key Laboratory of Cognitive Intelligence \\
$^3$Institute of Artificial Intelligence, Hefei Comprehensive National Science Centerce\\
$^4$iFLYTEK Research\\
\small \texttt{\{doublecheng, ghzhao0223, guoyx18, linx\}@mail.ustc.edu.cn} \\ \small \texttt{\{huangzhy, leexin\}@ustc.edu.cn}  \\ 
\small \texttt{\{jzwu4, sjwang3\}@ifytek.com} 
}

\begin{document}
\maketitle
\begin{abstract}
Automatically generating high-quality mathematical problems that align with educational objectives is a crucial task in NLP-based educational technology.  Traditional generation methods focus primarily on textual quality, but they often overlook educational objectives. Moreover, these methods address only single-dimensional, simple question generation, failing to meet complex, multifaceted educational requirements. To address these challenges, we constructed and annotated EduMath, a dataset of 16k mathematical questions with multi-dimensional educational objectives. Based on this dataset, we developed EQGEVAL, which incorporates three evaluation dimensions and is designed to assess the ability of models to generate educational questions. Drawing inspiration from teachers' problem design processes, we propose the Educational Question Planning with self-Reflection (EQPR) method for educational mathematical question generation, following a "plan-evaluate-optimize" approach. Specifically, by combining planning algorithm based on Monte Carlo Tree Search with the generative capabilities of Large Language Models, we continuously optimize questions through iterative feedback. This self-optimization mechanism ensures that the generated questions both fit the educational context and strategically achieve specific basic educational objectives. Through extensive experiments based on EQGEVAL, we have demonstrated that EQPR achieves significant improvements in generating questions that meet multi-dimensional educational objectives.
\end{abstract}
\begin{figure}[t]
    \centering
   \vspace{0.6cm} 
    \includegraphics[width=0.9\linewidth]{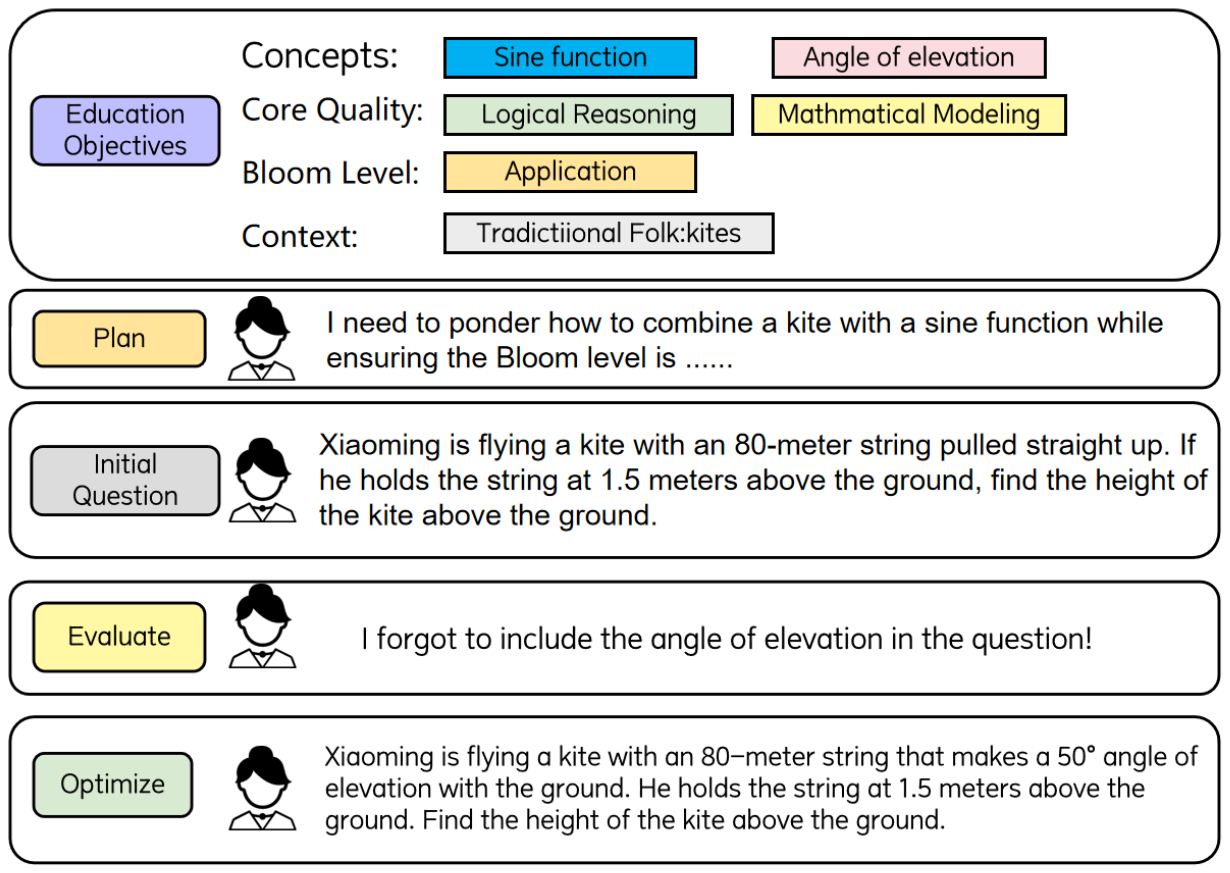} % Adjust width relative to linewidth
    \caption{A simple example of question design based on educational objectives}
\vspace{-0.6cm}
    \label{fig:case}
\end{figure}
\section{Introduction}
\label{intro}
Mathematical questions are fundamental elements of educational assessment and cognitive development, playing an irreplaceable role in cultivating students' logical thinking and problem-solving abilities ~\citep{kurdi2020systematic,liu2019ekt}. When designing such questions, educators coordinate multiple educational objectives(e.g., concepts, mathematical qualities ), not only to accurately evaluate students' degree of knowledge mastery and application skills but also to enhance their comprehensive problem-solving capabilities through progressive challenges. 
Figure \ref{fig:case} illustrates the systematic process of how teachers design test questions ~\citep{kliebard1970tyler, wiggins2005understanding}. Starting with educational objectives, teachers carefully consider multiple dimensions including core concepts, competencies to be assessed, and real-world scenarios. They strategically combine these elements to create engaging questions that connect theoretical knowledge with practical applications. Throughout this process, teachers continuously evaluate whether their questions adequately cover all intended objectives and make necessary refinements.

However, in the literature, the generation of mathematics questions that consider multiple educational objectives has not received sufficient attention. In widely studied mathematical question generation, previous work has relied solely on manually crafted templates for generation~\citep{polozov2015personalized,khodeir2018generating}, unable to generate content based on educational objectives expressed in natural language. Conversely, most more advanced works employ seq2seq models for generation~\citep{zhou2019towards,liu2020mathematical}, but due to model limitations, their applications remain confined to question generaton based on mathematical formulas such as arithmetic operations and equation systems. With the emergence of large language models like ChatGPT~\citep{kojima2022large} and LLaMA~\citep{dubey2024llama}, which significantly enhance the ability to generate diverse and complex content, researchers have begun using few-shot learning or Chain-of-Thought templates for question generation, framing it as a goal-based reasoning task. However, question generation is a complex task, and single-step reasoning may lead to failures, as seen in the teacher's first attempt in Figure \ref{fig:case}. Consequently, the limitations of single-step reasoning—its inherent inability to handle complex problems requiring multi-step inferences and its lack of reflective capabilities to check for errors—significantly hinder the full potential of large language models in question generation.

To validate the use of large language models for creating educational questions, we empirically assess whether these models can match human teachers' ability to design problems that achieve specific educational objectives. However, existing mathematics question generaton datasets present significant limitations. Most datasets focus primarily on elementary-level mathematics (e.g., LMWP~\citep{liu2020mathematical}, HMWP~\citep{qin2020semantically}), offering a narrow scope of assessment. Furthermore, a lack of comprehensive annotation is prevalent; the majority of these datasets (e.g., Gaokao-bench~\citep{zhang2023evaluating}, GSM8K~\citep{cobbe2021training}) lack annotations for educational objectives, hindering a thorough evaluation of LLMs' capabilities in mathematical question generation.

In this paper, to advance the field further, we present two datasets, EduMath-SQ and EduMath-CQ, which are filtered from real high school test papers and annotated with educational objectives. Specifically, for each question, we annotate four to five categories of educational objectives based on Tyler's rationale~\citep{kliebard1970tyler} and two-way specification table~\citep{odiagbe2016table}, aiming to comprehensively evaluate models' objective-based mathematical question generation capabilities.

Along this direction, several technical challenges remain. First, there is a lack of evaluation metrics. Past question generation work has adopted text quality metrics such as BLEU and ROUGE. For instance, in Figure \ref{fig:case}, while the initial and modified questions show high textual similarity, the initial question lacks a concept compared to the latter. However, these traditional text generation evaluation metrics often fail to effectively measure mathematical question generation quality, particularly in terms of educational objective alignment. Second, questions with multidimensional educational objectives typically cannot be generated in a single attempt,  as the complexity of multiple objectives often leads to certain objectives being overlooked or missed in a single generation attempt,  requiring repeatedly evaluated and optimized by educators. Additionally, a single dimension of an educational objective may encompass multiple components—as illustrated in Figure~\ref{fig:case}, where the "concepts" dimension includes both sine functions and elevation angles—further increasing the challenge of accurately generating questions that satisfy multidimensional educational objectives.

To address these challenges, we first propose EQGEVAL, a novel evaluation benchmark based on the EduMath dataset that comprehensively evaluates question generation quality through LLM using three metrics: solvability (question feasibility), Pass Rate (objective fulfillment), and Win Rate (comparison with gold standards). Furthermore, in real-world educational settings, teachers typically design mathematical questions through an iterative process, repeatedly evaluating and optimizing questions based on educational objectives and student feedback. Inspired by this practice, we develop a methodology that mirrors this systematic refinement process. Specifically, we develop the \textbf{E}ducational \textbf{Q}uestion \textbf{P}lanning with self-\textbf{R}eflection (EQPR) method, which combines Monte Carlo Tree Search (MCTS) with large language models to simulate diverse question design strategies, achieve educational objectives, and systematically evaluate and refine the question creation process through continuous feedback.

In summary, our contributions are:
\begin{itemize}
    \item We introduce a comprehensive dataset derived from authentic educational assessments, together with EQGEVAL, a novel benchmark framework designed to evaluate large language models' proficiency in generating questions that are precisely aligned with specified educational objectives.
    \item Our framework, EQPR, integrates Monte Carlo Tree Search (MCTS) for strategic planning with a reflection mechanism that evaluates question quality and proposes improvements, enabling iterative refinement of the generated questions.
    \item Empirical validation through extensive experimentation underscores our framework's efficacy. When implemented with two distinct LLMs and evaluated on the EQGEVAL benchmark, EQPR achieved superior performance across nearly all evaluation metrics, establishing new state-of-the-art results.
\end{itemize}

\section{Related Works}
\subsection{Question Generation}
Question generation is a significant research direction in educational technology~\citep{kurdi2020systematic,zhao2024diffusion,zhao2024comprehensive}. Its core function lies in automatically generating questions from structured or natural language text, such as deriving questions from dialogue content~\citep{guo2024pcqpr,liu2024socraticlm} or extracting them from story texts~\citep{li2024planning}. The key value of this technology is that it can substantially reduce the time and cost of manual design and construction of questions, while also dynamically generating questions of varying difficulty and type based on the content.
\begin{figure*}[t]
    \centering
    \vspace{-12pt}
    \includegraphics[width=\linewidth]{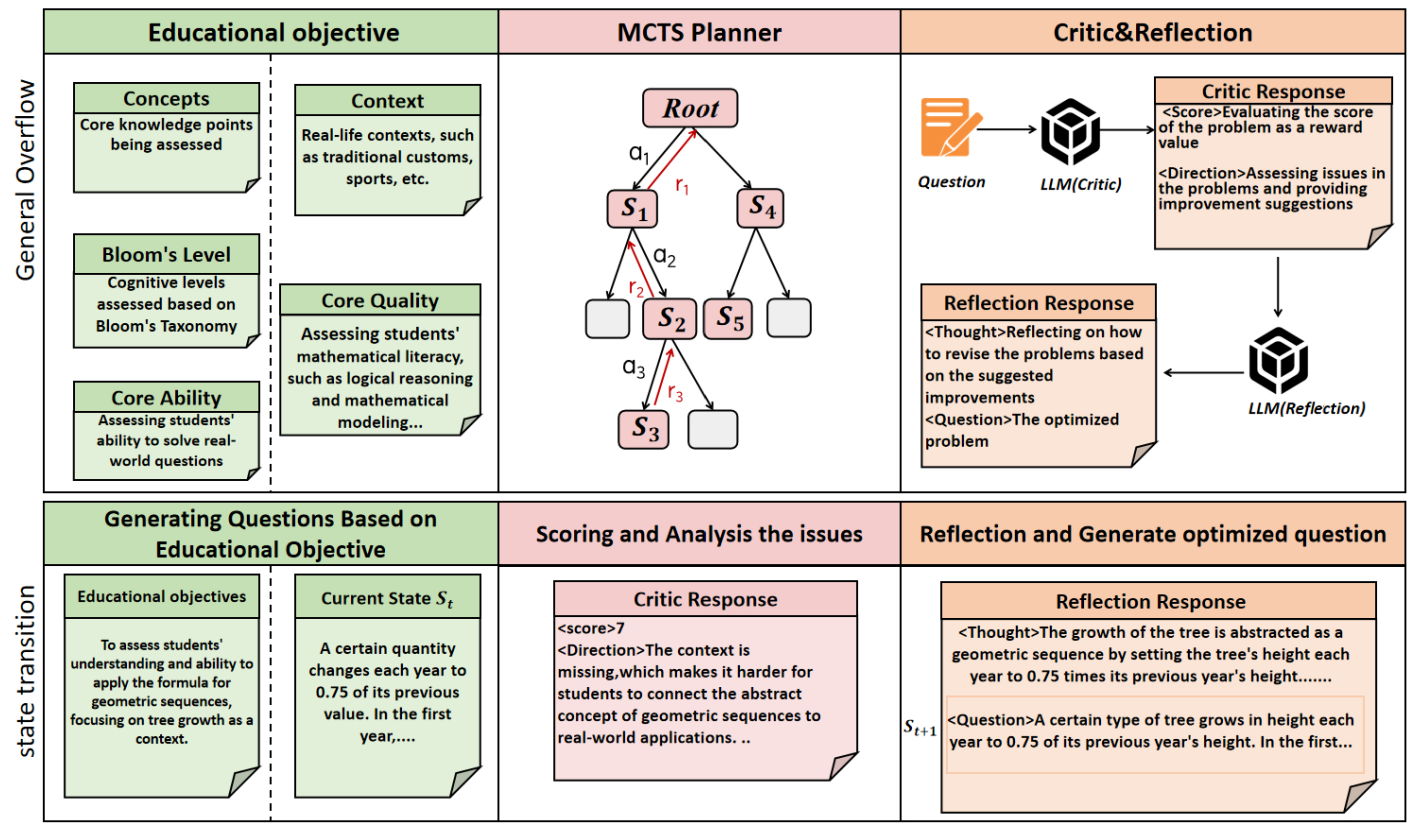} % Adjust width relative to linewidth
    \caption{The overall framework of EQPR.The upper part of the figure illustrates the entire EQPR process, which uses an MCTS structure for deep reasoning and iteratively improves question quality via the Critic and Reflection modules. The lower part shows a simplified state transition example, where a previous question is modified based on feedback and progresses to the next state.}
    \vspace{-12pt}
    \label{fig:model}
\end{figure*}

In mathematics education, question generation has been a significant area of research. Early work focused on generating mathematical questions based on mathematical formulas and scenarios, relying on predefined templates and rule-based methods. Researchers developed language knowledge bases and rhetorical structure rules to aid in question generation~\citep{khodeir2018generating}. As natural language processing evolved, the field shifted toward neural network-based approaches. Sequence-to-sequence frameworks for mathematical question generation~\citep{zhou2019towards} integrated thematic and formulaic information using attention mechanisms. Building on pre-trained language models, later work improved topic word selection and introduced question-solving modules to enhance the solvability of generated questions~\citep{wang2021math}. MapKG~\citep{qin2023mathematical}, inspired by educational experts' test design experience, advanced the field with a "plan-then-generate" strategy that incorporated dual attention mechanisms and knowledge graphs to improve question solvability and diversity. Recent developments in large language models have opened new research avenues, with methods such as gradient-based techniques being used to generate questions with controlled difficulty levels~\citep{lin2024idgen}. However, existing studies have largely overlooked the generation of questions that address multi-dimensional educational objectives. Our proposed EQPR model aims to fill this gap.  

\subsection{LLM Planning and Reflection}
Recent research on planning with large language models (LLMs) has witnessed remarkable advancement, significantly impacting domains such as common-sense reasoning~\citep{zhu2023dyval,xuedecompose} and embodied intelligence~\citep{sun2024adaplanner}. A central focus has been enhancing LLMs' capacity for systematic, step-by-step problem solving. Foundational techniques like Chain-of-Thought (CoT) prompting~\citep{wei2022chain} have established the paradigm of incremental problem decomposition, while more sophisticated approaches such as Tree-of-Thoughts (ToT)\citep{yao2024tree} and Reasoning via Planning (RAP)\citep{hao2023reasoning} explore solution spaces through hierarchical tree structures, leveraging methodologies like Monte Carlo Tree Search to systematically expand the search space. However, extensive iterative planning processes can introduce cumulative error propagation. To address this limitation, Self-Refine~\citep{madaan2024self} incorporates iterative feedback mechanisms that enable models to reflect upon and refine their reasoning processes, thereby mitigating error accumulation. Distinguished from these planning-focused methodologies, our work introduces EQPR, which synergistically combines iterative planning with reflective mechanisms to optimize question generation processes.

\section{Problem Formulation}

As outlined in Section \ref{intro}, our objective is to leverage large language models for generating questions that align with specified educational objectives \(O\), encompassing fundamental goals (e.g., conceptual understanding, Bloom's taxonomy levels) and their contextual frameworks (e.g., traditional cultural narratives, sports scenarios), formalized as \(q = LLM(O)\).

However, single-pass question generation often yields limited quality and inadequate alignment with educational objectives. To address this limitation, we draw inspiration from teachers' iterative improvement processes, modeling this as a sequential decision-making problem and formalizing it as a Markov Decision Process (MDP).

We formalize this question improvement process as an MDP defined by the tuple $(\mathcal{S}, \mathcal{A}, \mathcal{T}, R)$, where:

\begin{itemize}
    \item \textbf{State space} $\mathcal{S}$: Each state $s_t \in \mathcal{S}$ represents the current version of an educational question at time step $t$, denoted as $q_t$. States may include metadata such as alignment with educational objectives $O$, revision history, or evaluative metrics.
    
    \item \textbf{Action space} $\mathcal{A}$: Each action $a_t \in \mathcal{A}$ corresponds to an improvement operation targeting specific dimensions of the question, such as ``enhancing conceptual clarity'' or ``increasing real-world relevance.'' As shown in the upper part of Figure~\ref{fig:model}, these actions are formulated as structured feedback that guides iterative improvement, offering detailed analysis of current shortcomings paired with actionable improvement strategies.
    
    \item \textbf{Reward function} $R: \mathcal{S} \times \mathcal{A} \rightarrow \mathbb{R}$: After executing action $a_t$ in state $s_t$, the system receives a reward based on the revised question's quality. This reward reflects multiple educational criteria including conceptual clarity, cognitive depth, and contextual appropriateness, assessed by a \textit{critic module} providing both numerical scores and formative feedback (detailed in Section~\ref{subsec:critic}).
    
    \item \textbf{Transition function} $\mathcal{T}: \mathcal{S} \times \mathcal{A} \rightarrow \mathcal{S}$: The system transitions to a new state $s_{t+1}$ after applying action $a_t$ to the current question $s_t$. This process is implemented by a \textit{reflection module}, which interprets the critic's feedback and generates an improved question accordingly (detailed in Section~\ref{subsec:ref}).
\end{itemize}

As illustrated in Figure~\ref{fig:model}, the process operates as follows: Given a current state $s_t$ and educational objectives $O$, the critic module analyzes the question and samples an improvement action $a_t \sim \textit{Critic}(a \mid s_t, O)$. Simultaneously, it assigns a reward $r_t$ and identifies specific improvement areas. This feedback is passed to the reflection module, which translates it into actionable revision instructions and generates an enhanced question better aligned with educational objectives. For instance, if the action is to ``increase cognitive depth,'' the reflection module may introduce more abstract or higher-order reasoning components. The newly generated question becomes the next state $s_{t+1}$, and the process continues iteratively until optimal educational alignment is achieved.

\section{Method}
\subsection{Overview}
Drawing inspiration from educators' systematic approach to question development~\citep{wang2022towards} - which encompasses planning, writing, evaluating, and optimizing - we introduce an innovative framework for generating educational questions that ensures both quality and alignment with educational objectives. The framework consists of three modules: the Critic module, which evaluates each generated question (state) based on multiple dimensions of educational objectives and provides directions for modification; the Reflection module, which analyzes the feedback from the Critic to determine optimization directions and refine the question generation process, leading to the creation of new questions; and the MCTS-based Planning module, directed by the Critic and Reflection modules, which provide the necessary guidance and constraints to enable it to systematically and efficiently navigate the vast and multifaceted search space, thereby exploring a wide range of potential question structures and formulations.

\subsection{Critic}
\label{subsec:critic}
We employ the Critic module to systematically evaluate the alignment between generated questions and educational objectives. Through Large Language Models (LLMs) equipped with Critic prompts (see Table \ref{tab:critic} ) and Chain-of-Thought reasoning, the module performs comprehensive assessment across multiple dimensions. Specifically, for a state $s_t$ at phase t and educational objective $O$, the Critic generates question modification directions $a_t$ and  $score_t$ as follows:
\begin{equation}
(score_t, a_t) = Critic(s_t, O),
\end{equation}
Here, $score_t$ represents the quantitative evaluation score, while $a_t$ denotes the suggested modification directions. The evaluation examines concept coverage, contextual relevance, conceptual coherence, etc., ensuring comprehensive assessment of educational requirements, and logical interconnections. This structured evaluation framework enables precise identification of gaps between generated questions and desired educational objectives, facilitating targeted improvements in subsequent iterations.

\subsection{Reflection}

Cognitive science research demonstrates that humans continuously iterate and reflect upon their thinking based on new information~\citep{frederick2005cognitive}, allowing them to correct errors and deepen understanding. This reflective process is particularly evident in educational settings, where teachers systematically refine questions through iterative evaluation and improvement to align with educational objectives.

Consider the example illustrated in the bottom part of  Figure \ref{fig:model}, where the initial question generaton lacks contextual materials. Through reflection, this limitation is identified, leading to an improved iteration that incorporates a practical tree-planting scenario, making the mathematical concept more accessible and applicable.

To formalize this reflective iteration process, we introduce a Reflection module powered by large language models. This module employs specialized reflection prompts (detailed in the Table \ref{tab:reflection}) to analyze the historical trajectory of question states and improvement suggestions, ensuring alignment with educational objectives. In the iteration step $t$, we construct the historical trajectory $\tau$ as:
\begin{equation}
\tau = \{(s_0, a_0), (s_1, a_1), \ldots, (s_{t-1}, a_{t-1})\},
\end{equation}
where \( s_0 \) represents the initial question state generated based on the initial generation prompt (see Table \ref{tab:init}) at the root node, $a_0$ denotes the Critic module's modification suggestions for initial question $s_0$, $s_t$ indicates the preceding state of the current node, and $a_t$ represents the Critic module's modification suggestions for question $s_t$. This trajectory captures the complete history of question states and their corresponding improvement suggestions up to the current iteration.
The next state generation is accomplished through the Reflection module:
\begin{equation}
s_{t+1} = Reflection(O, \tau, s_t, a_t),
\end{equation}
where $O$ represents the educational objectives.

This formalized reflection mechanism enables the question generation process to emulate human teachers' approach, continuously iterating to enhance problem quality.
\label{subsec:ref}
\subsection{MCTS-based Planning}
We combine large language models (LLMs) with Monte Carlo Tree Search (MCTS) to better meet our educational objectives. MCTS efficiently explores the question-generation space, balancing exploration and exploitation to improve the educational value of the resulting questions. As shown in the top panel of Figure \ref{fig:model}, we model the process as a search tree: each node represents a partially generated question, and each edge denotes an editing action—such as extending or rewriting the text. The overall planning loop follows the standard MCTS phases of selection, expansion, simulation, and backpropagation, detailed in Algorithm \ref{alg:mcts_reflection} in the Appendix.

\noindent\textbf{Selection. } The selection phase chooses the most promising node from the tree's branches for further exploration. Starting from the root node $s_0$ (initial question), the selection phase iteratively chooses the most promising nodes for exploration. To balance between known high-quality question structures (exploitation) and exploring new directions for improvement (exploration), we use the well-known Upper Confidence Bounds applied to Trees (UCT)~\citep{kocsis2006bandit} for node selection, as shown below:
\begin{equation}
\label{uct}
UCT(s_t, a_t) = Q(s_t, a_t) + c \sqrt{\tfrac{\ln N(s_t)}{N(ch(s_t, a_t))}},
\end{equation}
\noindent where $Q(s_t, a_t)$ is the potential future reward of applying action $a_t$ at time $t$, $N(s_t)$ is the number of visits to node $s_t$, $ch(s_t, a_t)$ is the child node reached after executing action $a_t$ in state $s_t$, and $c$ is a constant used to adjust exploration. At each level of the tree, the child node with the highest UCT value is selected.

\noindent\textbf{Expansion. }During the expansion phase, we utilize the Reflection module to generate new question states. This module leverages the historical optimization trajectory $\tau$ to analyze patterns from previous modifications, generating new candidate questions according to: $s_{t+1} = Reflection(O, \tau, s_t, a_t)$. To explore a wider range of question designs, multiple nodes are generated as child nodes in each expansion step.

\noindent \textbf{Simulation. } From the expanded nodes, the simulation phase explores potential question optimization paths via simulations. In each simulation, the system evaluates options based on estimated cumulative rewards, selecting the highest-reward path for further exploration. This continues until a terminal state, yielding a comprehensive simulation of potentially effective question optimization schemes.

\noindent \textbf{Backpropagation. } When a simulation reaches a terminal state, backpropagation begins, using the cumulative future reward obtained at the terminal node (illustrated by the red arrows in the central MCTS tree section of Figure \ref{fig:model}) to update the Q-values of each state-action pair, with the aim of refining future question selection.

\section{Educational QG Dataset \& Benchmark}
\label{bench}

Existing mathematical question generaton datasets primarily focus on elementary-level content and often lack comprehensive educational annotations. To address these limitations, we developed EduMath, aiming to create a high-quality dataset with annotation accuracy exceeding 90\% across all educational dimensions.
We developed this multi-dimensional annotation framework based on the following considerations: concept mapping reflects knowledge coverage, ability assessment captures problem-solving requirements, Bloom's Taxonomy represents cognitive levels, mathematical literacy evaluation aligns with educational objectives, and real-world context identification demonstrates practical value. These dimensions characterize the educational attributes of mathematical questions from distinct perspectives. We sourced 16k high-school level mathematics problems from mock exams and college entrance examinations, implementing strict quality filters to exclude problems with images or incomplete solutions.
Using DeepSeek-V3, we conduct an iterative three-round annotation process. Initial annotations are reviewed for accuracy and consistency. Specifically, we employ Chain-of-Thought prompting to guide multiple large language models in evaluating annotation correctness through a voting mechanism. Annotations flagged as inaccurate are re-annotated to resolve identified issues. This rigorous, multi-stage process results in an annotation accuracy of 95.2\% across all dimensions. The final dataset includes two variants: EduMath-SQ (Standard Questions) and EduMath-CQ (Contextual Questions), with EduMath-SQ focusing exclusively on non-contextual problems.

Building on this, we define controllable educational question generation and corresponding evaluation metrics. Previous question generation approaches typically rely on text quality metrics such as BLEU and ROUGE for evaluation, but these are far from sufficient. Truly effective questions must be solvable and meet teachers' instructional and assessment needs. Following TOOLEVAL~\citep{qin2023toolllm} and based on DeepSeek-V3, we propose EQGEVAL, which includes the following metrics (see Appendix \ref{emd} for details): Solvability as a fundamental requirement that the generated problems must have valid solutions; Pass Rate measuring the proportion of generated problems that meet the educational objectives; and Win Rate where we present educational objectives and two problems to DeepSeek-V3 evaluators, asking them to determine which problem better serves the intended purpose. To ensure the reliability of our evaluation, all metrics are determined through a majority voting mechanism, where multiple independent evaluations are conducted to derive the final assessment results.

\begin{table*}[!t]
\small
\centering
\caption{Evaluation results on datasets EduMath-CQ and EduMath-SQ(\%).}
\renewcommand{\arraystretch}{1.1}
\setlength{\tabcolsep}{6pt}
\resizebox{\textwidth}{!}{%
\begin{tabular}{l l ccccccc}
\toprule
\rowcolor{gray!10}
\textbf{Dataset} & \textbf{Method} & \textbf{BLEU} & \textbf{METEOR} & \textbf{ROUGE-L} & \textbf{BERTSCORE} & \textbf{WIN RATE} & \textbf{SOLVABLE} & \textbf{PASS RATE} \\ 
\midrule
\multirow{11}{*}{\centering \textbf{EduMath-CQ}}
& \multicolumn{8}{c}{\cellcolor{gray!10}\textbf{GPT-4o-Mini}} \\
& COT     & 11.60  & 29.64 & 24.45  & 68.52 & 35.78 & 90.12 & 27.88 \\
& COT-BON & 8.52   & 22.59 & 10.56  & 69.19 & 34.50 & 90.23 & 21.25 \\
& REACT   & 18.90  & \textbf{53.20} & 36.72  & \textbf{72.55} & 38.40 & 89.32 & 33.66 \\
& DEAR    & 13.31  & 23.09 & 21.74  & 70.34 & 35.18 & 85.77 & 27.35 \\
& EQPR    & \textbf{24.80} & 52.70 & \textbf{48.13} & 70.89 & \textbf{39.92} & \textbf{90.63} & \textbf{35.81} \\ 
\cmidrule(lr){2-9}
& \multicolumn{8}{c}{\cellcolor{gray!10}\textbf{DeepSeek-V3}} \\
& COT     & 12.11  & 42.90 & 22.51  & 70.64 & 38.20 & 90.84 & 31.45 \\
& COT-BON & 13.58  & 43.60 & 28.93  & 70.85 & 38.99 & 90.25 & 29.63 \\
& REACT   & \textbf{21.76}  & 47.98 & \textbf{47.06} & 70.32 & 39.75 & 90.64 & 38.91 \\
& DEAR    & 20.71  & \textbf{49.06} & 42.11  & \textbf{74.48} & 41.81 & \textbf{92.59} & 37.88 \\
& EQPR    & 20.33  & 46.86 & 44.57  & 71.02 & \textbf{46.23} & 91.73 & \textbf{43.11} \\
\midrule
\multirow{11}{*}{\centering \textbf{EduMath-SQ}}
& \multicolumn{8}{c}{\cellcolor{gray!10}\textbf{GPT-4o-Mini}} \\
& COT     & 1.76   & 21.04 & 7.64   & 67.30 & 32.20 & 83.65 & 64.60 \\
& COT-BON & 2.63   & 30.44 & 11.76  & 68.43 & 36.64 & \textbf{85.23} & 75.12 \\
& REACT   & \textbf{36.72} & 52.19 & 36.72  & \textbf{72.55} & 36.81 & 82.90 & 83.22 \\
& DEAR    & 8.51   & 20.52 & 4.71   & 66.34 & 36.70 & 81.89 & 82.36 \\
& EQPR    & 32.77  & \textbf{52.75} & \textbf{52.78} & 68.83 & \textbf{37.18} & 84.40 & \textbf{84.37} \\ 
\cmidrule(lr){2-9}
& \multicolumn{8}{c}{\cellcolor{gray!10}\textbf{DeepSeek-V3}} \\
& COT     & 28.80  & 56.91 & 50.33  & 75.38 & 37.40 & 87.15 & 85.46 \\
& COT-BON & \textbf{30.66} & \textbf{61.50} & \textbf{52.07} &  \textbf{76.05}  & 42.70 & 86.88 & 88.82 \\
& REACT   & 26.46  & 60.47 & 46.05  & 75.45 & 44.45 & 85.95 & 89.76 \\
& DEAR    & 24.52  & 58.23 & 40.07  & 73.48 & 40.36 & 87.14 & 82.74 \\
& EQPR    & 29.71  & 57.29 & 49.63  & 74.79 & \textbf{45.65} & \textbf{91.31}  & \textbf{91.50} \\
\bottomrule
\end{tabular}%
}
\label{tab:merged_results_v2}
\end{table*}
\section{Experiment}
\subsection{Experimental Settings}
\vpara{Datasets.} As outlined in Section \ref{bench}, we conducted experimental evaluations using the EduMath-SQ and EduMath-CQ datasets. We randomly selected 10\% of the data to serve as the test set. The EduMath-CQ dataset comprises 589 educational objectives paired with their corresponding gold-standard questions, while EduMath-SQ contains 1,034 educational objectives along with their respective gold-standard questions. For more detailed information about these datasets, please refer to Appendix \ref{appx:data}.

\vpara{Baselines.} Since this is a novel task, we approach it as a reasoning problem and benchmark it against established reasoning methods, including Chain-of-Thought (CoT) \citep{wei2022chain}, CoT-BON, ReAct \citep{yaoreact}, and the tree-structured DEAR method \cite{xuedecompose}. DeepSeek-V3 \cite{deepseekv32024Deepseek} and GPT-4o-Mini \citep{gpt4omini2024openai} serve as our primary backbone models. Additionally, we evaluated Claude-3.5 \citep{sonnet2024anthropic} and GPT-4o \citep{gpt4o2024openai} on the EduMath-CQ dataset; for those results, please refer to Appendix \ref{AdditionalResults}.

\vpara{Metrics.} We evaluate our approach using both automatic metrics and human assessment. For automatic evaluation, we primarily employ three metrics from the EQGEVAL framework, where the Win Rate metric compares generated questions against gold-standard problems. To comprehensively assess text quality, we also incorporate widely-adopted natural language generation metrics, including Rouge-L ~\citep{rouge}, BLEU ~\citep{bleu}, METEOR~\citep{meteor}, and BERTSCORE~\citep{bertscore}. For human evaluation, we design two metrics: Fluency and Human-rated Win Rate. The Fluency metric uses a three-level scoring system (0-2), with detailed scoring criteria provided in the Appendix \ref{criteria}.

\subsection{Main Results}
We conduct experiments to verify the effectiveness of our framework EQPR, and report the results in Table \ref{tab:merged_results_v2}.
We get the following observations. Primarily, our method demonstrates superior performance in Win Rate and Pass Rate across both EduMath-CQ and EduMath-SQ datasets, as well as across both GPT-4o-Mini and DeepSeek models. This comprehensive outperformance validates the effectiveness of our approach. Notably, when testing with DeepSeek on EduMath-CQ, our method achieved a Win Rate of 46.23\%, surpassing the next-best method DEAR (41.8\%) by a substantial margin of 4.42\%. This improvement clearly demonstrates the efficacy of our iterative refinement strategy. Nevertheless, we observed that Pass Rates consistently decreased when transitioning from standard questions (EduMath-SQ) to contextual scenarios (EduMath-CQ), suggesting that large language models still face challenges in seamlessly incorporating contextual elements while addressing multiple educational objectives.

Regarding Solvability, our method achieves near-optimal performance, ranking second across both sub-datasets. The slight reduction in solvability scores can be attributed to the nature of our iterative refinement process, where questions naturally evolve to become more sophisticated through multiple iterations.

In terms of traditional text quality metrics, our method maintains competitive performance while prioritizing educational effectiveness. For instance, in the EduMath-SQ dataset with DeepSeek, the modest difference between our method's BERTScore (74.79) and REACT's (76.05) is acceptable, considering that these metrics primarily assess lexical and semantic similarities rather than educational value. Our superior Win Rate and Pass Rate scores underscore our method's success in achieving its primary objective: generating educationally meaningful and high quality questions, even if this occasionally leads to slightly lower linguistic metric scores.

\subsection{Human Evaluation Results}
\begin{table}[t]
\centering

\setlength{\tabcolsep}{4pt} % 稍微减小全局列间距
\newcolumntype{?}{!{\vrule width 1pt}}
\renewcommand\arraystretch{1.0}
\scalebox{0.9}{
\begin{tabular}{@{}c@{ }?@{\hspace{1em}}cc@{}} % 针对两列指标缩小间距
\toprule
\textbf{Method} & \textbf{Clarity} & \textbf{Win Rate-Human(\%)} \\
\midrule
COT & 1.83 & 27.00 \\
COT-BON & 1.78 & 33.67 \\
DEAR & 1.84 & 31.00 \\
REACT & 1.87 & 35.00 \\
EQPR & \textbf{1.93} & \textbf{36.67} \\
\midrule
Fleiss' kappa & 0.71 & 0.47 \\
\bottomrule
\end{tabular}
}
\caption{Human evaluation results on EduMath-CQ(DeepSeek-V3).}
\label{tab:my_human_evaluation}
\end{table}
To thoroughly assess the effectiveness of our methodology, we carried out human evaluation experiments. We randomly selected 100 samples from the EduMath-CQ dataset and enlisted three highly educated evaluators, each with at least a bachelor's degree, to conduct the assessments. The evaluation focused on two key dimensions: question clarity (whether the questions are easily readable and understandable) and quality comparison. Given the potential biases in large language models when determining win rates, the evaluators were asked to compare the quality of the generated questions against the gold-standard questions. As depicted in Table \ref{tab:my_human_evaluation}, our method demonstrated superior performance in both clarity scores and human-evaluated win rates. Furthermore, we evaluated the reliability of the annotations using Fleiss' Kappa coefficient. The Kappa values for both clarity and quality comparison exceeded the credibility threshold of 0.41, thereby confirming the reliability of our evaluation results.
\subsection{Ablation Results}
\begin{figure}[t]
    \centering
    \vspace{-8pt}
    \includegraphics[width=\linewidth]{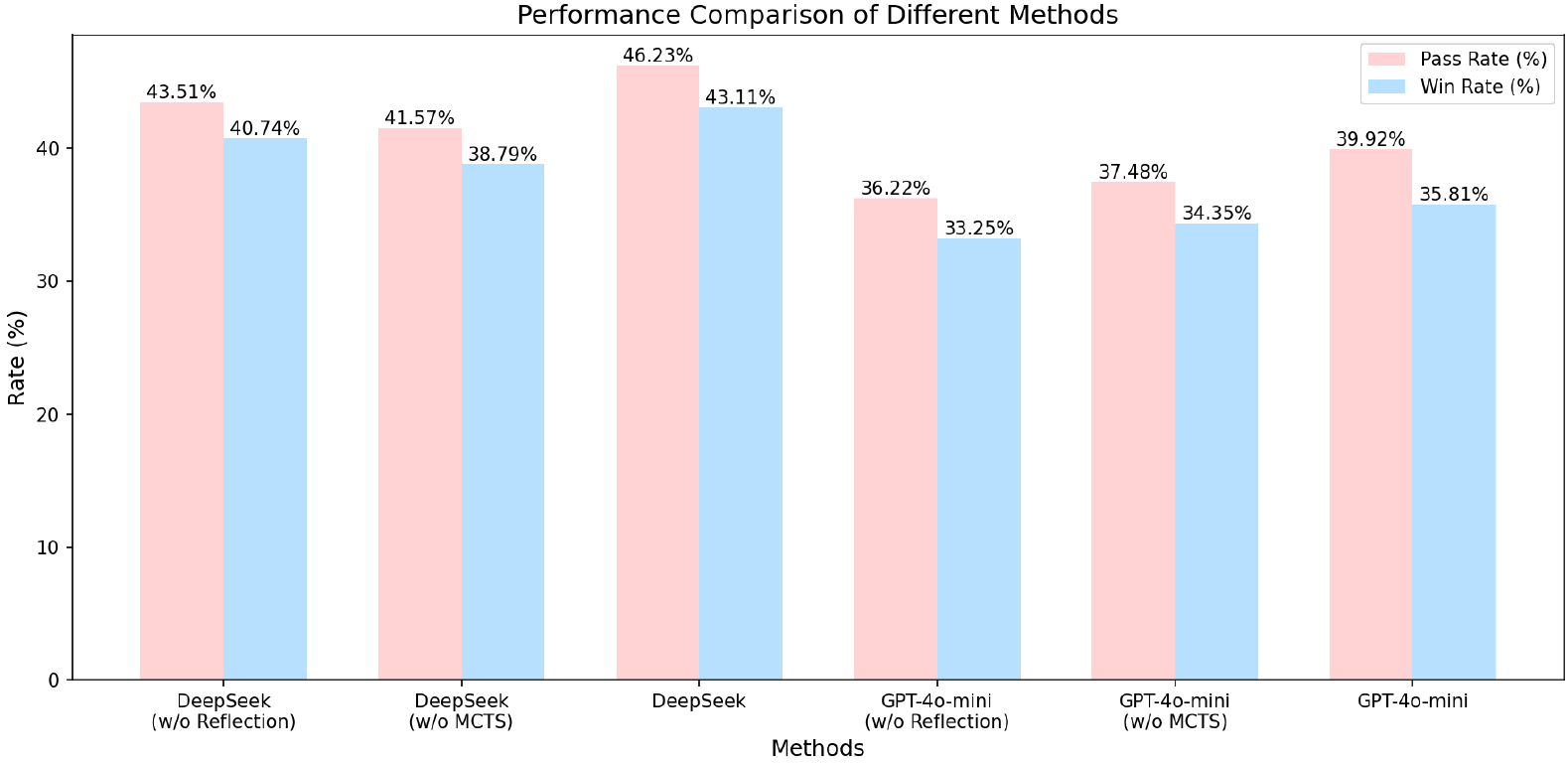} 
    \caption{The results of ablation studies. We test different methods on EduMath-CQ datasets}
    \vspace{-8pt}
    \label{fig:abltion}
\end{figure}

\vpara{Effect of Reflection Module.}
To validate the effectiveness of the Reflection Module, we conducted a comparative experiment. After removing the Reflection Module, the model no longer optimizes based on iterative feedback but directly generates questions according to the target (referred to as the "w/o Reflection" experiment). The results showed significant performance degradation across both models: using DeepSeek as an example, illustrated in Figure \ref{fig:abltion}, the Pass Rate decreased from 46.23\% to 43.51\%, and the Win Rate declined from 43.11\% to 40.74\%. These observations highlight the vital importance of the Reflection Module. By engaging in multiple rounds of iterative optimization, the system adeptly integrates knowledge from previously encountered questions, enabling it to produce content that more effectively aligns with educational objectives.

\vpara{Effect of MCTS-based Planning.}To quantify the contribution of MCTS-based planning within our proposed framework, we conducted an ablation study by removing the MCTS component and employing greedy search exclusively for action selection. As demonstrated in Figure \ref{fig:abltion}, the model's performance deteriorated substantially without MCTS-based planning. Specifically, DeepSeek experienced a notable decline in both pass rate (from 46.23\% to 41.57\%) and win rate (from 43.11\% to 38.79\%) when operating without Monte Carlo Tree Search. We attribute this performance degradation to MCTS-based planning's superior capability in navigating the question optimization space through its dual mechanism of prospective outcome prediction and retrospective evaluation. These empirical findings validate the critical role of MCTS integration in our framework's effectiveness.

\section{Conclusion}
In this paper, we introduced EQPR (Educational Question Planning with self-Reflection), an innovative framework for generating high-quality mathematical questions that align with educational objectives. EQPR integrates a "plan-evaluate-optimize" process, combining Monte Carlo Tree Search with the generative power of LLMs, enabling continuous refinement through feedback optimization. We also introduced EduMath, a high-quality dataset of 16k mathematics problems, and EQGEVAL, a comprehensive framework for evaluating the educational value of generated questions. Extensive experiments demonstrate that EQPR outperforms existing reasoning methods on key educational metrics across multiple large language models.
\section*{Acknowledgement}
This research was partially supported by the National Science and Technology Major Project(No.2022ZD0117103), the National Natural Science Foundation of China (Grants No.62477044), Anhui Provincial Natural Science Foundation (No. 2308085QF229), the Fundamental Research Funds for the Central Universities (No.WK2150110038). Zhenya Huang gratefully acknowledges the support of the Young Elite Scientists Sponsorship Program by CAST (No. 2024QNRC001)
\section*{Limitations}
Our research primarily focused on mathematical question generation and has not yet been extended to other subject areas. This presents an important direction for future research. Additionally, we face certain challenges in establishing educational objectives, particularly in the assessment of question difficulty. Since difficulty evaluation is largely subjective and challenging to standardize, this remains a significant hurdle in the field that requires further investigation. Furthermore, while we utilize large language models for evaluation, these models may exhibit certain biases, and their assessment results do not always align perfectly with the professional judgment of human educators. Consequently, exploring effective methods to align human evaluation with LLM-based assessments and establishing a more accurate evaluation system remains a crucial direction for future research.
\bibliography{main}

\appendix
\section{More Experiment Details}
\subsection{Dataset Details}
\label{appx:data}

As shown in the Table \ref{tab:stats}, we present the number of questions for each dataset along with the average number of concepts covered per question. On average, each question involves more than two concepts, indicating that we intend to use these datasets to evaluate the ability of large language models to generate questions that address complex educational objectives. To this end, we randomly sampled 10\% of the questions from each dataset to serve as a test set.
\begin{figure*}[!t]
    \centering
    \vspace{-8pt}
    \includegraphics[width=\linewidth]{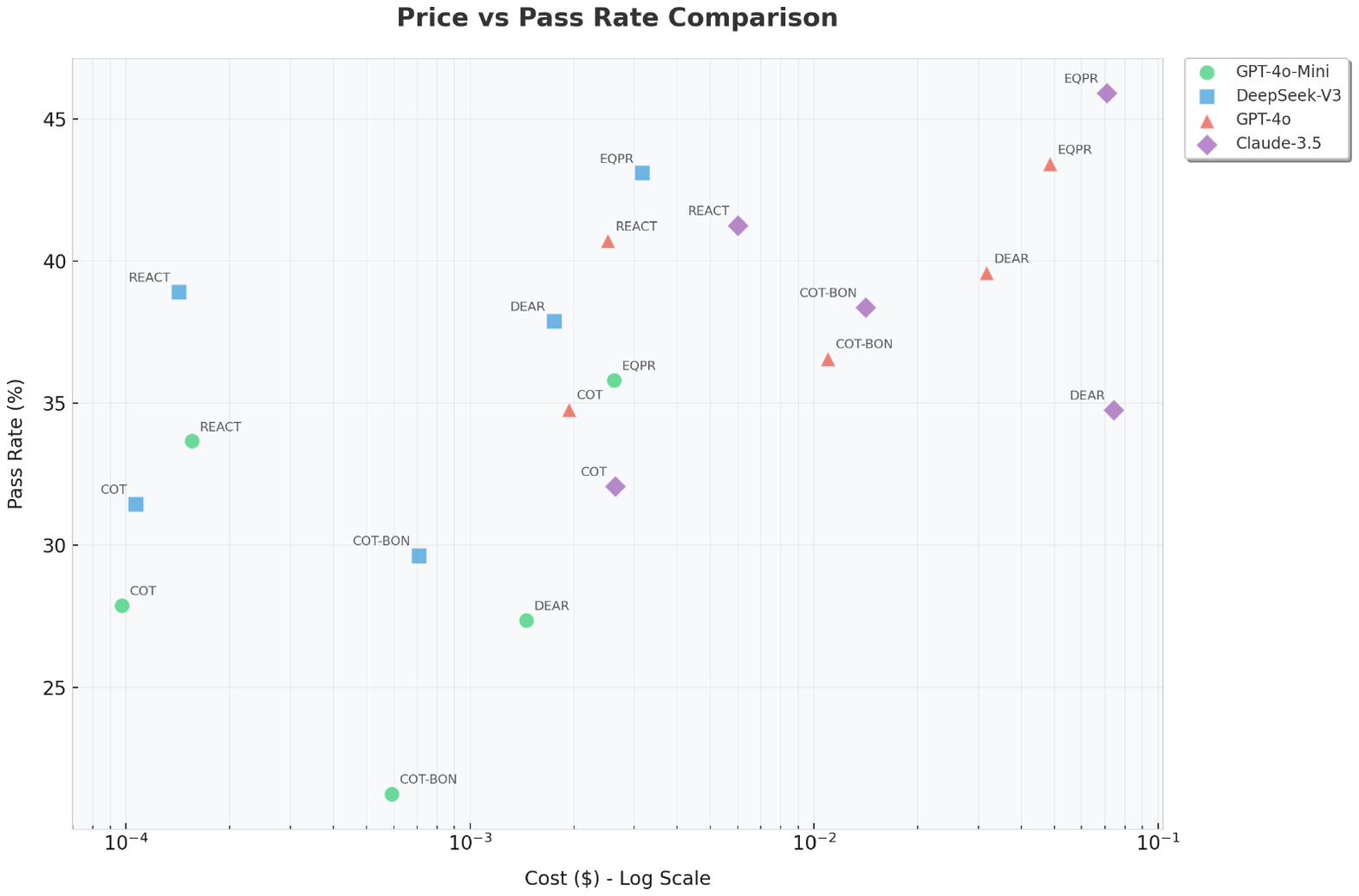} 
    \caption{The cost refers to the average cost required to generate a question on the EduMath-CQ dataset.}
    \vspace{-8pt}
    \label{fig:cost}
\end{figure*}

\subsection{Explanation of the Educational Objectives}
\begin{itemize}
\item \textbf{Concept:} Understand and master core mathematical concepts such as trigonometric functions, sequences, and probability.
\item \textbf{Core Quality:} Develop essential skills such as logical reasoning, mathematical modeling, and problem-solving to tackle complex mathematical tasks.
\item \textbf{Core Ability:} Develop the ability to choose and apply appropriate mathematical techniques—for instance, recognizing when to use identities like \( \sin^2x + \cos^2x = 1 \) in solving problems.

\item \textbf{Bloom Level:} Design questions that intentionally target specific cognitive levels in Bloom’s Taxonomy, such as application, analysis, or creation.
\item \textbf{Context:} Enable students to interpret and solve mathematical problems within real-world scenarios and authentic contexts.

\end{itemize}

\subsection{Dataset Comparation}
We conducted a comprehensive comparison between our dataset and existing mathematics question generaton datasets. As shown in Table \ref{tab:compara}, Our EduMath dataset demonstrates several significant advantages. First, the scale of our dataset substantially surpasses other comparable datasets in terms of problem quantity. Second, our dataset specifically focuses on high school mathematics problems that require deeper cognitive reasoning, whereas datasets like LMWP and HMWP primarily target elementary-level mathematics question generaton. The generation of high school mathematics problems presents considerably greater challenges due to their complexity and reasoning requirements.
Furthermore, our dataset features meticulous annotations performed by large language models (LLMs). We implemented a rigorous two-stage annotation process: after the initial annotation phase, we employed LLMs to verify the accuracy of the annotations, with any identified inaccuracies undergoing a re-annotation process. This makes EduMath the first and only open-source high school mathematics dataset that incorporates multi-dimensional educational objective annotations, setting a new standard for educational resource development.
\begin{table*}[t]
\small  
\centering
\setlength{\tabcolsep}{4pt}  
\caption{Comparison of Different Mathematics Problem Datasets}
\begin{tabular}{l|c|c|c|c|c}
\hline
\multirow{2}{*}{Datasets} & EduMath & GAOKAO & LMWP & HMWP & GSM8k \\[-0.5ex]
 & (Ours) & (\citet{zhang2023evaluating}) & (\citet{liu2020mathematical}) & (\citet{qin2020semantically}) & (\citet{cobbe2021training}) \\
\hline
Deep Reasoning & \checkmark & \checkmark & \ding{55} & \ding{55}  & \ding{55} \\
Objectives Annotation & \checkmark & \ding{55} & \checkmark & \checkmark & \ding{55} \\
Multi-Edu objectives & \checkmark & \ding{55} & \ding{55} & \ding{55} & \ding{55} \\

\hdashline
\#Problems & 16348 & 300 & 5447 & 5470 & 8500 \\

\hline
\end{tabular}
\label{tab:compara}
\end{table*}
\subsection{Educational Metric Details}
\label{emd}
\vpara{Solvable}
We employed the state-of-the-art large language model, DeepSeek-V3, to solve the generated questions. We then determined the solvability of each question using a majority voting approach based on self-consistency (five samples).

\vpara{Pass Rate}
We employ Chain of Thought (COT) reasoning by inputting both educational objectives and generated questions into the large language model, enabling step-by-step analysis of whether the question meets each educational objective. Using the Self-Consistency approach, we generate 5 independent judgment samples and determine the final result through majority voting. If an question fails to meet any educational objective, it is marked as failing.

\vpara{Win Rate}The Win Rate metric evaluates question quality by inputting pairs of questions along with their educational objectives into the large language model to determine which is superior. The evaluation criteria include adherence to educational objectives, natural language flow, and seamless integration of context. As shown in Table \ref{tab:education_objectives_evaluation}, we provide a specific case study demonstrating this judgment process.

\subsection{Implementation Details}
\label{appx:llm}
\vpara{Parameter Details.}For LLM parameter settings, we maintained a consistent temperature of 0.7 across all four models to ensure output diversity. The model versions used in our testing include gpt-4o-mini-2024-07-18, Claude-sonnet-3.5-0622, GPT-4o-2024-11-20, and DeepSeek-V3. Regarding method parameters, for COT-BOT (Best of N), we generated 5 candidate outputs and selected the best result; for our proposed EPQR method, the Monte Carlo Tree Search (MCTS) parameters were configured as follows: 4 iterations, maximum depth of 3, and an exploration parameter c of 2.5 in the UCT formula.

\vpara{Output strategy.}Each iteration of the Monte Carlo Tree Search (MCTS) yields a path from the root node to a leaf node. Following PromptAgent~\citep{wang2023promptagent}, we select the path with the highest average question reward and then choose the question with the highest reward from that path as the final output. This strategy ensures that we identify the best question from the overall optimal search trajectory.
\begin{table}[t]
\small
  \centering
  \caption{Number of Questions and Average Number of Concepts for Two Datasets}
  \label{tab:stats}
  \begin{tabular}{l r r}
    \toprule
    \textbf{Dataset} & \textbf{\# Questions} & \textbf{Avg. \# Concepts} \\
    \midrule
    EduMath-SQ & 10763 & 2.57 \\[2pt]
    EduMath-CQ & 5585 & 2.39 \\
    \bottomrule
  \end{tabular}
\end{table}
\subsection{Baseline Details}
\label{baseline}
\begin{itemize}
\item \textbf{Chain-of-thoughts (CoT)~\citep{wei2022chain}:} Prompts language models to think step-by-step before reaching final conclusions, incorporating deliberate reasoning and systematic thinking to generate more powerful and insightful answers.

\item \textbf{Chain-of-thoughts Best of N ~\citep{wei2022chain}:} Samples multiple CoT outputs and selects the best one from the generated candidates.
\item \textbf{ReAct~\citep{yaoreact}:} Simulates human problem-solving patterns through reasoning and action steps, enabling large language models to better understand tasks, gather information, execute operations, and correct errors, thereby significantly improving their performance on complex tasks. For the education question generation task, we adopt a thought + action approach, where the model generates a thought before producing the final question.
\item \textbf{Dear~\citep{xuedecompose}:} A human cognition-inspired reasoning framework that builds a reasoning tree through a three-stage cycle. It decomposes complex problems into sub-problems in the Decompose stage, generates and self-checks reasoning processes for each sub-problem in the Analyze stage, and updates parent node reasoning based on child node results in the Rethink stage, thereby enhancing large language models' complex reasoning capabilities.

\end{itemize}

\subsection{Human Evaluation Criteria}\label{criteria}
\begin{itemize}
\item 0: Incomprehensible - The question is confusing and impossible for students to understand, making it impossible to answer.
\item 1: Partially Clear - Students can grasp the core idea of the question and attempt to answer, but the question still needs improvement.
\item 2: Completely Clear - The question is concise, clear, easy to understand, and allows students to answer smoothly.
\end{itemize}

\section{Additional Results}\label{AdditionalResults}
\subsection{Evaluation of Methods on More LLMs}
\begin{table}[!t]

\footnotesize  % Make text smaller
\centering
\caption{Additional Evaluation results on datasets EduMath-CQ(\%).}
\renewcommand{\arraystretch}{1.1}  % Adjust row spacing
\setlength{\tabcolsep}{4pt}  % Adjust column spacing (smaller value)
\begin{tabular}{l|cc}
\toprule
\rowcolor{gray!10}
\textbf{Method} & \textbf{WIN RATE} & \textbf{PASS RATE} \\ 
\midrule
\multicolumn{3}{c}{\cellcolor{gray!5}\textbf{GPT-4o}} \\
\midrule
COT     & 38.16& 34.78 \\
COT-BON & 34.34 & 36.58 \\
REACT   & 42.68 & 40.72 \\
DEAR    & 41.30 & 39.59\\
Ours    & \cellcolor{blue!10}\textbf{44.07} & \cellcolor{blue!10}\textbf{43.43} \\
\midrule
\multicolumn{3}{c}{\cellcolor{gray!5}\textbf{Claude-3.5}} \\
\midrule
COT     & 37.95 & 32.08 \\
COT-BON & 41.61 & 38.37 \\
REACT   & 45.27 & 41.26 \\
DEAR    & 44.19 & 34.75 \\
Ours    & \cellcolor{blue!10}\textbf{47.11} & \cellcolor{blue!10}\textbf{45.92} \\

\bottomrule
\end{tabular}%
\label{tab:add}
\end{table}

To validate the model-agnostic nature of our framework, we conducted experiments on EduMath-CQ using GPT-4o and Claude-3.5 as backbone models, with results presented in Table \ref{tab:add}. The experimental results demonstrate that our framework consistently outperforms the baseline across both models in terms of Win Rate and Pass Rate metrics. This consistent superior performance indicates that our framework's effectiveness is model-agnostic and can be successfully applied across various large language models. Additionally, Claude-3.5 demonstrates the strongest performance among the four models, likely attributed to its enhanced reasoning capabilities and superior instruction-following abilities. This finding further indicates that the performance of the foundation model has a significant impact on the overall effectiveness of the framework.
\begin{figure}[!t]
    \centering
   
    \includegraphics[width=\linewidth]{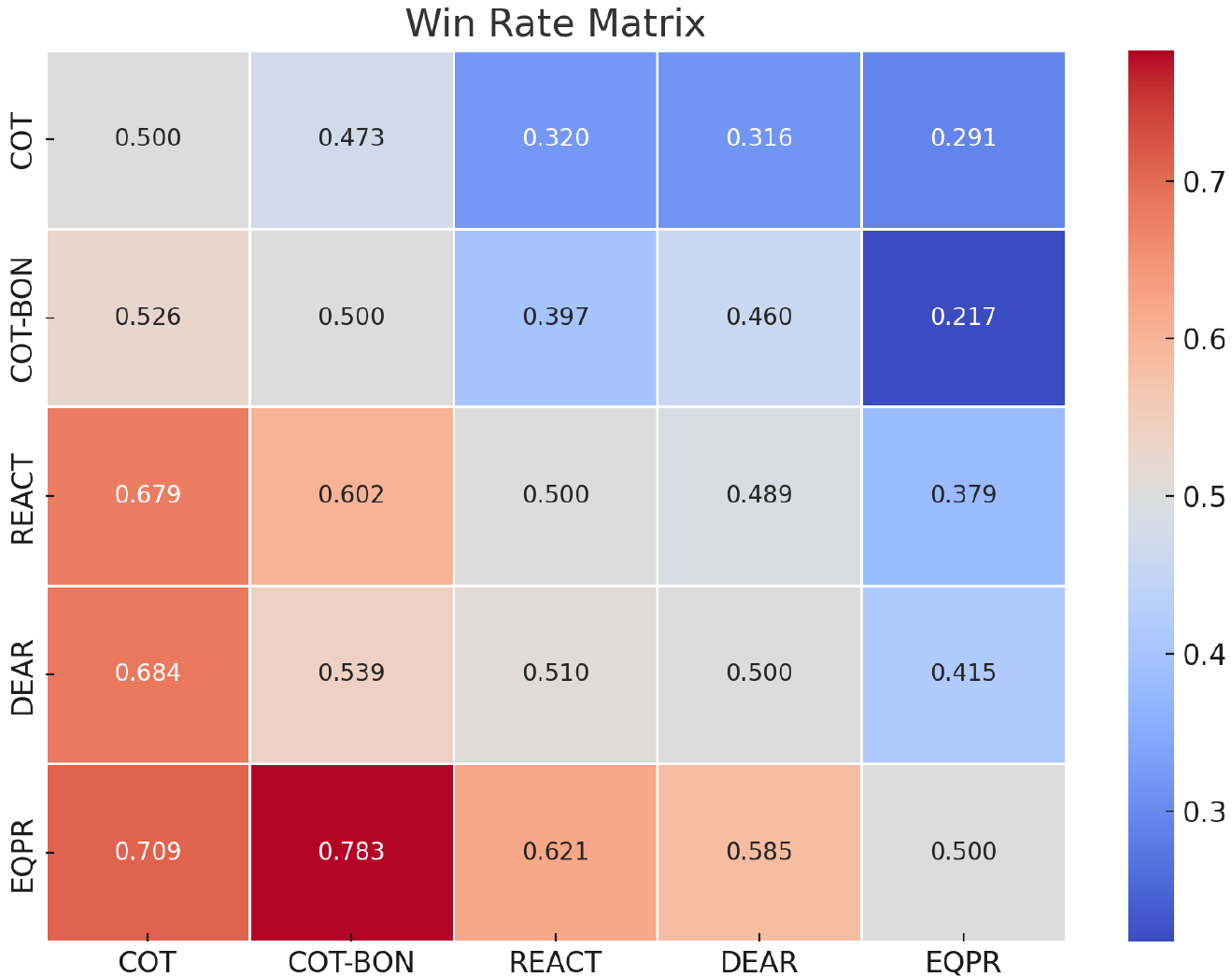} % Adjust width to 70% of linewidth
    \caption{Win Rate Comparison Matrix Across Different Methods}
   
    \label{fig:win}
\end{figure}

\subsection{Cost Analysis}
We presented a comparative analysis of four models on the EduMath-CQ dataset, examining both the average cost per question generation and the corresponding pass rates. The Figure \ref{fig:cost} demonstrates that all models performed better under the EQPR method compared to the baseline. Notably, DeepSeek achieved exceptional cost-effectiveness, attaining the third-highest pass rate at less than \$0.01 per question, significantly lower than the costs of Claude-3.5 and GPT-4o. While the Claude-3.5-based EQPR solution had the highest cost, it achieved the best pass rate, with expenses still remaining under \$0.1 per question. Given that the cost of human-generated questions would be considerably higher than this amount, we consider this investment to be 
justified.
\subsection{Win Rate Comparison of Methods}
To evaluate the performance of these five methods in a more fine-grained manner, we conducted pairwise comparisons on the EduMath-CQ dataset based on DeepSeek-V3's results and calculated their respective Win Rates, as detailed in the Figure \ref{fig:win}. The results demonstrate that our proposed method, EQPR, outperforms other methods in terms of Win Rate, highlighting its superiority. Specifically, EQPR exhibits significant advantages even when compared to strong baseline methods. For instance, EQPR achieved Win Rates of 58.5\% and 62.1\% against DEAR and REACT, respectively. This suggests that EQPR is more effective in guiding the model to generate high-quality questions that are both coherent and aligned with educational objectives.
\begin{table*}[t]
\small
\centering
\caption{Case study of an example generated by our method (Claude-3.5)}
\label{tab:random_event_evaluation}
\begin{tabularx}{0.95\textwidth}{lX}
\toprule
\multicolumn{2}{c}{\textbf{Education Objectives}} \\
\midrule
\textbf{Education Objectives:} &
\begin{itemize}[nosep, leftmargin=*]
    \item \textbf{Concept:} Random events
    \item \textbf{Core Quality:} Mathematical abstraction, logical reasoning
    \item \textbf{Core Ability:} Distinguish between random, certain, and impossible events, and make probability judgments in real‐life contexts
    \item \textbf{Context:} Real‐life scenarios involving random events (e.g., dice rolling, lottery, baby gender) to enhance relevance and interest while testing key concepts
    \item \textbf{Bloom's Taxonomy:} Understanding
\end{itemize} \\
\midrule
\multicolumn{2}{c}{\textbf{Question}} \\
\midrule
\textbf{Question:} &
\begin{minipage}[t]{0.7\textwidth}
Which of the following describes a \emph{random} event?\\[2pt]
A.\;It will rain tonight\quad
B.\;The sun will rise in the west tomorrow\quad
C.\;There are 12 months in a year\quad
D.\;A coin toss will definitely land heads up
\end{minipage} \\
\midrule
\multicolumn{2}{c}{\textbf{Updated}} \\
\midrule
\textbf{Critic:} &
\begin{minipage}[t]{0.7\textwidth}
\textbf{score:} 6.\\[4pt]
\textbf{reason:} Covers basic concept, but lacks real‐life context; omits “impossible event”; low cognitive demand; insufficient instructional depth\\
\end{minipage} \\

\midrule
\textbf{Reflection:} &
\begin{minipage}[t]{0.7\textwidth}
\textbf{thought:}Include random, certain, and impossible options with plausible distractors; keep single‐choice format and moderate difficulty; use parallel comparisons to guide classification by definition.\\[4pt]
\textbf{question:} Which of the following describes a \emph{random} event?\\
A.\;Rolling a fair die shows a 6.\\
B.\;The sun will rise in the west tomorrow.\\
C.\;There are 12 months in a year.\\
D.\;A tossed coin will definitely land heads up.\\
\end{minipage} \\
\bottomrule
\end{tabularx}
\vspace{-10pt}
\end{table*}

\subsection{case study}
To better illustrate the process of question generation, critic, and revision, Table \ref{tab:random_event_evaluation} presents a case study based on the learning objective of "categorizing random events." The system first generates an initial question based on predefined education objectives : “Which of the following describes a random event?...” The Critic module then evaluates the question, assigning a score of 6 and identifying several issues, such as the lack of a real-world context, failure to address impossible events, and limited cognitive depth. In response, the Reflection module proposes concrete revisions—for example, ensuring that the answer choices include random, certain, and impossible events, maintaining a single-choice format, and keeping the difficulty level moderate. Based on these suggestions, an improved version of the question and answer options is generated. This case study demonstrates how the question quality can be progressively refined to better align with instructional goals and cognitive requirements.

\clearpage

\begin{algorithm*}[]
\caption{MCTS with Reflection}

\begin{algorithmic}[]
\small
\REQUIRE Initial question(state) $s$ (root), expand width $k$, exploration weight $w$, $max\_iterations$, $depth\_limit$, reflection module $reflection$, critic module $critic$

\STATE Initialize root node with initial state
\FOR{$i \gets 1$ \TO $max\_iterations$}
    \STATE $path \gets []$
    \STATE $node \gets root$

    \STATE \textcolor{blue}{\# Selection Phase}
    \WHILE{$node$ has children}
        \STATE $node \gets \argmax_{child \in node.children}(child.uct)$
        \STATE Add $node$ to $path$
        \STATE Update $node.visited$
    \ENDWHILE

    \STATE \textcolor{blue}{\# Expansion Phase}
    \IF{$node.depth < depth\_limit$}
        \FOR{$j \gets 1$ \TO $k$}
            \STATE Generate new child through reflection:
            \STATE $question \gets Reflection$
            \STATE $child \gets create\_node(question, parent=node)$
            \STATE Evaluate $child.reward$ using $Critic$
            \STATE Add $child$ to $node.children$
        \ENDFOR
    \ENDIF

    \STATE \textcolor{blue}{\# Simulation Phase}
    \WHILE{not terminal $node$}
        \STATE Select child with the highest immediate reward
        \STATE Add $node$ to $path$
        \STATE Update $node.visited$
    \ENDWHILE

    \STATE \textcolor{blue}{\# Backpropagation Phase}
    \STATE $cumulative\_reward \gets 0$
    \FOR{$node$ in $reversed(path)$}
        \STATE $\begin{multlined}
                cumulative\_reward \gets \\
                cumulative\_reward + node.reward
               \end{multlined}$
        \STATE $\begin{multlined}
                \text{Update } node.cum\_rewards \text{ with } \\
                cumulative\_reward
               \end{multlined}$
        \STATE Calculate new $node.Q$ value
    \ENDFOR

\ENDFOR
\RETURN the best question from the path

\end{algorithmic}
\label{alg:mcts_reflection}
\end{algorithm*}

\clearpage

\begin{table*}[h]

\centering
\begin{tabular}{|p{0.7\textwidth}|}
\hline
\textbf{Question Generation Prompt} \\
\hline
You are an expert in high school mathematics education. You are analyzing educational objectives to design and create a multiple-choice question. Your goal is to develop a well-structured question that aligns with specific educational objectives while fostering core competencies. \\
\textbf{Status Determination Rules:} \\
\textbf{concepts:} Ensure complete alignment with required content; Maintain logical rigor and clear progression; Consider cognitive levels \\
\textbf{Competency Development:} Integrate core competencies naturally with content; Build connections between concepts \\
\textbf{Example:}
\{few\_shots\}\\
\textbf{Output Format:} \\
\{"question\_design\_thought": "detailed explanation of question design approach", "question": "complete multiple-choice question with options"\} \\
\textbf{Required Input:} Education\_Objectives: \{educational objectives\} \\
\hline
\end{tabular}
\caption{Question generation prompt template.}
\label{tab:init}
\end{table*}

\begin{table*}[!h]

\centering
\begin{tabular}{|p{0.7\textwidth}|}
\hline
\textbf{Critic Evaluation Prompt} \\
\hline
You are an expert in high school mathematics education. Your task is to evaluate a mathematical question and its design approach based on given educational objectives. You will assess whether the question meets the educational objectives, provide a strict scoring evaluation, and analyze areas for improvement. \\
\textbf{Scoring Scale (1-10):} \\
\textbf{Excellent (10):}
- Complete alignment with Concept and competency requirements
- Clear, structured design following stated approach 
- Deep pedagogical design fostering core competencies
- Appropriate cognitive level for students \\
\textbf{Good (8-9):}
- Generally meets educational objectives
- Minor deviations from target cognitive level \\
\textbf{Average (6-7):}
- Meets content requirements but lacks competency development
- Cognitive level misalignment with objectives \\
\textbf{Fair (4-5):}
- Only partially meets content requirements  
- Significant cognitive level misalignment \\
\textbf{Poor (1-3):}
- Severe deviation from educational objectives
- Does not follow design approach
- Low quality question \\
\textbf{Failing (0):}
- No connection to educational objectives
- Completely unrelated to goals and design approach \\
\textbf{Key evaluation points:} \\
\textbf{Concept Alignment:}
- Check for complete coverage of required content
- Assess logical structure and progression
- Evaluate cognitive level appropriateness \\
\textbf{Competency Development:}
- Analyze integration of core competencies
- Assess effectiveness in building understanding \\
\textbf{Output Format:} \\
\{"direction": "detailed analysis of weaknesses, and improvement suggestions", "score": numerical score 1-10\} \\
\textbf{Required Input:} Education\_Objectives: \{educational objectives\}; Question: \{current question\} \\
\hline
\end{tabular}
\caption{Critic evaluation prompt template.}
\label{tab:critic}
\end{table*}

\begin{table*}[!h]

\centering
\begin{tabular}{|p{0.7\textwidth}|}
\hline
\textbf{Reflection Prompt} \\
\hline
You are an expert in high school mathematics education. Your task is to analyze and optimize a math question and its design approach based on given educational objectives, previous feedback, and the question's evolution history. Your goal is to refine both the question design approach and the question itself to better meet educational objectives while maintaining high quality. \\
\textbf{Key Constraints:} \\
- All questions must have valid solutions
- Modifications should go beyond numerical changes
- Changes must align with existing objectives
- No new concepts or competencies can be added \\
\textbf{Analysis Points:} \\
\textbf{Previous Question Analysis:}
- Review strengths and weaknesses identified
- Understand suggested improvements 
- Study evolution of previous versions \\
\textbf{Optimization Strategy:}
- Address identified weaknesses
- Maintain existing strengths
- Enhance alignment with objectives
- Improve question quality \\
\textbf{Educational Alignment:}
- Verify concept coverage
- Check competency development \\
\textbf{Output Format:} \\
\{"thought": "detailed explanation of optimized design approach", "question": "complete optimized question"\} \\
\textbf{Required Input:} 
Education\_Objectives: \{educational objectives\}
Question: \{current question\}
Direction: \{current feedback and suggestions\}
Trajectory\_thoughts: \{history of previous versions\} \\
\hline
\end{tabular}
\caption{Reflection prompt template.}
\label{tab:reflection}
\end{table*}

\begin{table*}[t]
\small
\centering
\caption{Example of the Input and Output for the Education Objectives Evaluation Module}
\label{tab:education_objectives_evaluation}
\begin{tabularx}{0.95\textwidth}{lX}
\toprule
\multicolumn{2}{c}{\textbf{Education Objectives}} \\
\midrule
\textbf{Education Objectives:} &
\begin{itemize}[nosep, leftmargin=*]
    \item \textbf{Concept:} Permutation and Combination, Counting Principle
    \item \textbf{Core Quality:} Mathematical Operations, Logical Reasoning
    \item \textbf{Context:} School Course Scheduling (combining combinatorial mathematics with practical school management)
    \item \textbf{Core Ability:}
    \begin{itemize}[nosep, leftmargin=1em] % Add some indentation for sub-items
        \item Handle complex permutation and combination problems.
        \item Apply classification counting and step-by-step counting techniques.
        \item Solve practical problems using enumeration and classification addition principles.
    \end{itemize}
    \item \textbf{Bloom's Taxonomy:} Application
\end{itemize} \\
\midrule
\multicolumn{2}{c}{\textbf{Question Pairs}} \\
\midrule
\textbf{Question 1:} & 
\begin{minipage}[t]{0.7\textwidth}
    A school has 3 courses to assign for the semester, and each course must be taught by one teacher. There are 2 available teachers to choose from for each course. How many possible ways can the teachers be assigned to the courses?

\end{minipage} \\
\midrule
\textbf{Question 2:} & 
\begin{minipage}[t]{0.7\textwidth}
    A school needs to arrange courses for the semester. There are 5 courses, and each course needs to be assigned a teacher. Assume each course can be taught by one of 3 teachers, and each teacher can teach at most two courses. How many ways can the courses be assigned to teachers?
\end{minipage} \\
\midrule
\multicolumn{2}{c}{\textbf{Evaluation}} \\
\midrule
\textbf{Better Question:} & 2 \\
\hdashline[1pt/2.5pt]
\textbf{Reason:} &  Question 1 only covers the basic counting principle of assigning teachers to courses, without considering any constraints. It does not address the permutation aspect, which is an important part of the core educational objectives. On the other hand, Question 2 is more complete as it covers both permutation and combination, as well as the counting principle, by introducing a constraint (each teacher can teach at most two courses). This makes Question 2 a better fit for evaluating the student’s ability to apply counting techniques and solve combinatorial problems. Therefore, Question 2 is a more suitable and well-rounded problem for this purpose.

 \\
\bottomrule
\end{tabularx}
\vspace{-10pt}
\end{table*}

\begin{table*}[!h]

\centering
\begin{tabular}{|p{0.7\textwidth}|}
\hline
\textbf{Pass Rate Evaluation Prompt} \\
\hline
Consider the following task as a mathematics education assessment assignment. You will act as a senior mathematics education expert, rigorously evaluating a given math problem and its associated educational objectives. You will assess based on the following six key dimensions, checking each item meticulously. If any item does not meet the requirements, the overall assessment result will be marked as a failure (outputting 0): \\
\textbf{Key Evaluation Dimensions:} \\
\textbf{1. concept Coverage:}
- Verify complete coverage of all required concepts
- Any omission or introduction of unmentioned points is non-compliant \\
\textbf{2. Bloom's Taxonomy Level:}
- Analyze cognitive level alignment with objectives
- Must neither exceed nor fall short of target level \\
\textbf{3. Student Problem-Solving Skills:}
- Confirm comprehensive assessment of all targeted skills \\
\textbf{4. Mathematical Core Competencies:}
- Evaluate cultivation of:
 * Logical reasoning
 * Mathematical operations
 * Spatial visualization
 * Data analysis
 * Mathematical modeling
 * Mathematical abstraction \\
\textbf{5. Rigor Requirement:}
- Maintain objectivity and rigor throughout evaluation
- Any non-compliance results in direct failure (0) \\
\textbf{Output Format:} \\
\{
 "reason": "Detailed explanation of the reasoning and process behind the evaluation",
 "pass\_rate": 1 or 0
\} \\
\textbf{Required Input:} 
Education\_Objectives: \{educational objectives\}
Question: \{question\} \\
\hline
\end{tabular}
\caption{Pass rate evaluation prompt template.}
\label{tab:pass_Rate}
\end{table*}

\begin{table*}[!h]

\centering
\begin{tabular}{|p{0.7\textwidth}|}
\hline
\textbf{Win Rate Evaluation Prompt} \\
\hline
As a senior mathematics education expert, please rigorously evaluate and compare the following question pair. In the evaluation process, analyze each question based on the following dimensions and determine which question better meets the educational objectives. \\
\textbf{Evaluation Dimensions:} \\
\textbf{1. Completeness of Concept Coverage:}
- Analyze coverage of required concepts
- Check for missing or redundant points \\
\textbf{2. Matching of Cognitive Levels:}
- Assess alignment with specified cognitive level
- Verify appropriate goal alignment \\
\textbf{3. Relevance to Ability Development:}
- Confirm effective training of specified abilities
- Verify alignment with outlined requirements \\
\textbf{4. Development of Mathematical Literacy:}
- Analyze contribution to mathematical literacy development \\
\textbf{5. Scientific Design of the Structure:}
- Evaluate reasonableness of question structure
- Assess organization and guidance quality \\
\textbf{6. Text Clarity and Coherence:}
- Assess clarity and conciseness of wording
- Evaluate effectiveness of problem-solving communication \\
\textbf{Output Format:} \\
\{
   "better\_question": 1 or 2,
   "reason": "Detailed evaluation reasons, explaining why the selected question is better and specifying which dimension(s) show superior performance."
\} \\
\textbf{Required Input:}
Education\_Objectives: \{educational objectives\}
Question Pair: \{Question pair\} \\
\hline
\end{tabular}
\caption{Win Rate evaluation prompt template.}
\label{tab:win_Rate}
\end{table*}

\end{document}